\documentclass{article}
\usepackage{styles/spconf}  
   
\usepackage{amsfonts}
\usepackage{array}
\usepackage[dvipsnames]{xcolor}

\usepackage{hyperref}
\usepackage{multirow}

\usepackage{graphicx}
\usepackage{subcaption}
\usepackage{booktabs}       
\usepackage{multirow}
\usepackage{mathtools}
\newcommand{\describe}[3][0pt]{\hspace*{.12em}\underbracket[0.5pt][1pt]{#2\hspace*{#1}}_\text{#3}}

\title{Towards Adversarial Robustness And Backdoor Mitigation in SSL}
\name{Aryan Satpathy$^{*1}$, Nilaksh$^{*1}$, Dhruva Abhijit Rajwade$^{*1}$, Somesh Kumar$^{1,2}$
\thanks{This work used the Supercomputing facility of IIT Kharagpur established under National
        Supercomputing Mission (NSM), Government of India and supported by Centre for Development of Advanced Computing (CDAC), Pune\\
        *Equally contributing authors\\
        $^1$:\{aryansatpathy, nilaksh404, rajwadedhruva\}@kgpian.iitkgp.ac.in, $^2$:smsh@maths.iitkgp.ac.in}}
\address{$^1$Indian Institute of Technology, Kharagpur\\
        $^2$Department of Mathematics, IIT Kharagpur}

\newcommand{\nil}[1]{}
\newcommand{\bonak}[1]{}

\begin{document}
\maketitle

\begin{abstract}
Self-Supervised Learning (SSL) has shown great promise  in learning representations from unlabeled data. The power of learning representations without the need for human annotations has made SSL a widely used technique in real-world problems. However, SSL methods have recently been shown to be vulnerable to backdoor attacks, where the learned model can be exploited by adversaries to manipulate the learned representations, either through tampering the training data distribution, or via modifying the model itself. This work aims to address defending against backdoor attacks in SSL, where the adversary has access to a realistic fraction of the SSL training data, and no access to the model. We use novel methods that are computationally efficient as well as generalizable across different problem settings. We also investigate the adversarial robustness of SSL models when trained with our method, and show insights into increased robustness in SSL via frequency domain augmentations. We demonstrate the effectiveness of our method on a variety of SSL benchmarks, and show that our method is able to mitigate backdoor attacks while maintaining high performance on downstream tasks. Code for our work is available at \href{https://github.com/Aryan-Satpathy/Backdoor}{https://github.com/Aryan-Satpathy/Backdoor}.
\end{abstract}

\section{Introduction}


\begin{figure}[tp]
    \centering
      \begin{minipage}[b]{\textwidth}
  {\includegraphics[width = 0.24\textwidth]{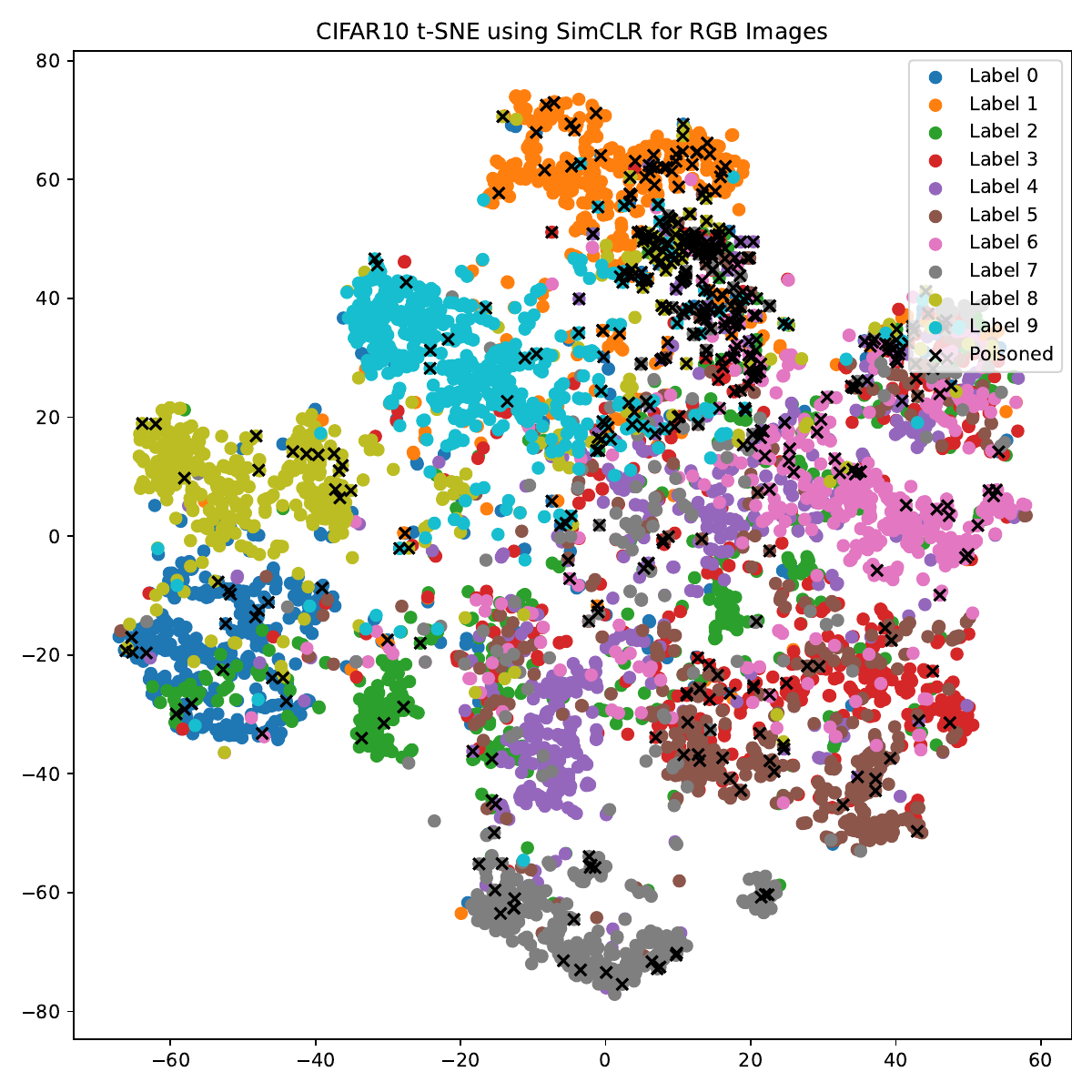}}
  {\includegraphics[width = 0.24\textwidth]{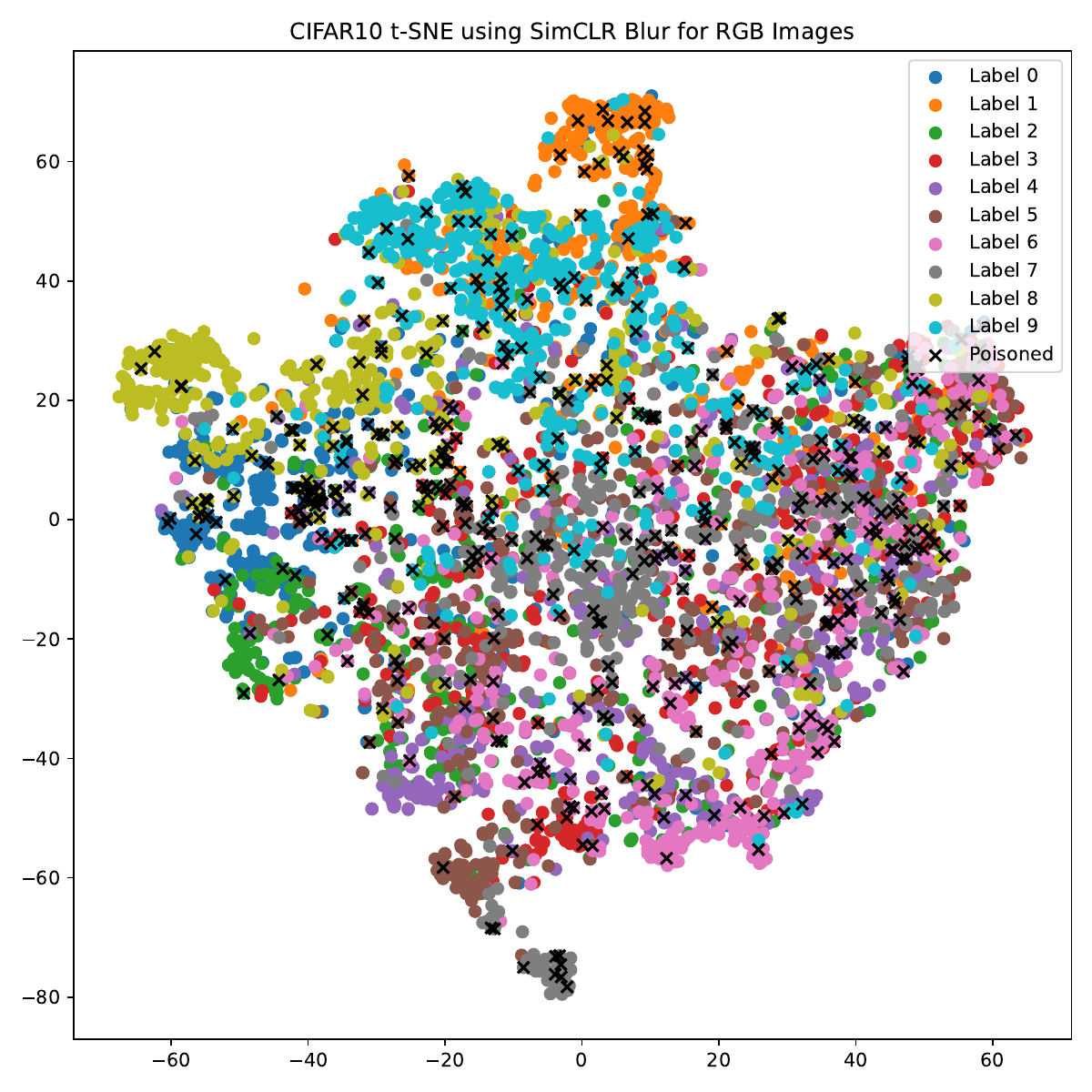}}  
  \end{minipage} 
    
    \caption{t-SNE Clustering of CIFAR10 embeddings obtained using SimCLR for a poisoned vanilla model (left) and a poisoned vanilla model equipped with the blur defense (right). While a poisoned sample \textbf{x} is clustered with the target class (orange) irrespective of its actual class for the poisoned model (left), the blur defense mitigates this and \textbf{x} is clustered with with its actual class (right).\\}
    \vspace{-0.7cm}
    \label{fig:clusters}
\end{figure}


Self-supervised learning (SSL) has become ubiquitous in many machine learning tasks, largely owing to the work of \cite{simclr},\cite{moco}, and \cite{simsiam}. Recently, \cite{CTRL} demonstrated a simple yet potent poison insertion strategy (CTRL) for SSL, by perturbing images in the \textit{frequency} space to poison a small fraction of the data (1\% of the entire training dataset). This poisoning method, being implemented in the frequency domain of the image, is invisible to the human eye and hard to detect via existing defense mechanisms. Additionally, this method is \textit{augmentation invariant}, making it highly effective against SSL. Some other backdoor attacks work through model access \cite{badencoder} or poisoning data subsets \cite{backdoor_attacks_umbc}. In this work, we explore the problem of defending against frequency-based backdoor attacks, in which a malicious entity plants “backdoors” into target models during training and activates such backdoors at inference. Various backdoor defense strategies aim to detect, recover, or mitigate poisoned samples. The first generalizable approach was introduced by \cite{neural_cleanse}, detecting poisoned samples via minimal perturbations and reverse-engineering the trigger to recover clean data. \cite{decoupling_backdoor} uses differential clustering on backdoored models, while \cite{smoothinv} regenerates poison patterns through a robust classifier. \cite{backdoor_defense_regularization} and \cite{distillationdefense}, \cite{distillationdefense2} apply knowledge distillation and weight initialization to defend against backdoors.

The problem of training SSL models to be robust to adversarial perturbations or leveraging adversarial effects for training models has become popular recently. We demonstrate our defense strategy against frequency-based attacks using CTRL but also theoretically show that our defense is partly generalizable to other types of attacks. We introduce two augmentations to increase model robustness for downstream classification tasks: the Blur augmentation, and a novel Frequency Patch augmentation that works in the frequency space. We also evaluate the adversarial robustness of our methods using the RobustBench \cite{robustbench} framework to demonstrate the improvement as a result of the augmentations.

\section{Preliminaries}

\subsection{Self-supervised and contrastive learning}

SSL learns representations by maximizing the similarity of embeddings for multiple views of the same input image $\mathbf{x}$. Formally, let $\mathcal{D}$ denote the dataset and $\mathcal{A}$ the set of possible augmentations. Mathematically, SSL aims to learn an encoder $f_\theta: \mathcal{X} \to \mathbb{R}^d$ parameterized by $\theta$, where $\mathcal{X}$ represents the input space and $d$ is the dimensionality of the embedding space. For a given image $\mathbf{x} \in \mathcal{X}$, the encoder $f_\theta$ outputs an embedding vector $\mathbf{e} = f_\theta(\mathbf{x}) \in \mathbb{R}^d$.

The training objective is to find the optimal parameters $\theta^*$ that maximize the similarity between the embeddings of augmented views of the same image, while minimizing the similarity with embeddings from different images:
\begin{equation*}
\begin{aligned}[t]
\theta^* = & \underset{\theta}{\operatorname{argmax}} \ \mathbb{E}_{\mathbf{x}, \mathbf{y} \sim \mathcal{D}}\mathbb{E}_{\mathbf{x}_a, \mathbf{x}_b \sim \mathcal{A}(\mathbf{x}), \mathbf{y}_a \sim \mathcal{A}(\mathbf{y})} \\
& \left[ - \operatorname{sim}\left(f_\theta(\mathbf{x}_a), f_\theta(\mathbf{y}_a)\right) + \operatorname{sim}\left(f_\theta(\mathbf{x}_a), f_\theta(\mathbf{x}_b)\right) \right]
\end{aligned}
\end{equation*}
where, $\mathbf{x}_a, \mathbf{x}_b \in \mathcal{X}$ are augmented versions of the input image $\mathbf{x}$, and $\mathbf{y}_a \in \mathcal{X}$ is an augmented version of a different image $\mathbf{y} \in \mathcal{D}$. The function $\operatorname{sim}:\mathbb{R}^d \times \mathbb{R}^d \to \mathbb{R}$ represents a similarity metric between two vectors. Two widely used similarity $\operatorname{sim}$ functions are cosine similarity given by $\operatorname{sim}_{\text{cos}}(\mathbf{e}_1, \mathbf{e}_2) = \frac{\mathbf{e}_1^\top \mathbf{e}_2}{\|\mathbf{e}_1\| \|\mathbf{e}_2\|}$ \cite{byol} and the normalized temperature-scaled cross-entropy loss (NT-Xent) \cite{simclr}.

\subsection{Equivariance and invariance}
\label{subsec:equi_inv}

Inspired from \cite{robustequivariance}, we use \emph{Equivariance} and \emph{Invariance} to improve robustness against adversarial and backdoor attacks at training time. A function $f$ is invariant or equivariant if it satisfies their respective conditions:
\begin{gather}
    \begin{aligned}
    &\text{Equivariance:} &&f(x) = g^{-1} \circ f \circ g(x)\\
    &\text{Invariance:} &&f(x) = f \circ g(x)
    \label{eq:inv-equi}
    \end{aligned}
\end{gather}
where $\circ$ denotes composition of functions, $g: \mathbb{S}\rightarrow\mathbb{S}$ is an arbitrary transformation (such that domain and co-domain are the same sets), and $g^{-1}$ is its inverse.

In computer vision, downstream tasks such as object detection or image segmentation are equivariant to augmentations, i.e., applying a rotation $r(\cdot)$ on an input image $x$ will also rotate the label ($r\circ f(x) = f\circ r(x)$). Whereas, classification tasks are largely\footnotemark{} invariant to augmentations, i.e., applying a rotation $r(\cdot)$ on $x$ will not change the class of the image ($f(x) = f\circ r(x)$).

By forcing a neural net $f_\theta$ to be invariant or equivariant, we enforce constraints on the perturbation signal (or trigger pattern) to be invariant or equivariant, respectively, w.r.t. the augmentations to successfully attack. In CTRL \cite{CTRL}, the authors identify this effect and choose a trigger pattern that is repetitive and symmetric (therefore invariant to many augmentations such as random crop and rotation).

\footnotetext{There are exceptions such as rotation in MNIST dataset can convert an image of digit 6 into 9.}


\subsection{Adversarial Perturbations}
\label{adversarial_perturb}
Let $x \in \mathbb{R}^d$ be an input point and $y \in \{1, \ldots, C\}$ be its correct label. For a classifier $f : \mathbb{R}^d \rightarrow \mathbb{R}^C$, we define a \textit{successful adversarial perturbation} with respect to the perturbation set $\Delta \subseteq \mathbb{R}^d$ as a vector $\delta \in \mathbb{R}^d$ such that
\[
\arg\max_{c \in \{1, \ldots, C\}} f(x + \delta)_c \neq y \quad \text{and} \quad \delta \in \Delta, \tag{1}
\]
where typically the perturbation set $\Delta$ is chosen such that all points in $x + \delta$ have $y$ as their label.

\nil{should we scrape this altogether?}\bonak{yes, scrapping luma for now}

\section{Method}
\label{sec:method}
\bonak{have to rewrite so that now we are talking about both freqPatch and blur, and both backdoor and adversarial}

\textit{Attacker threat model:} Following the threat model from \cite{backdoor_attacks_umbc} and \cite{CTRL}, we assume that the attacker can pollute a small fraction of the training dataset. The attacker has no information about the training regime or network architecture the victim uses.\\
\textit{Defender assumptions:} We assume that the defender has full control over the training procedure. The defender doesn't require access to a small subset of the clean dataset, which is often required for existing defenses based on knowledge distillation, e.g., \cite{distillationdefense}, \cite{distillationdefense2}.\\

Since contrastive learning is already used in SSL, we must carefully craft and select augments to maximize robustness against attacks. In CTRL, the authors demonstrate that injecting triggers in the frequency domain results in a symmetric and repetitive pattern in image space, which renders their attack successful against contrastive learning. To this end, we add two frequency domain augmentations, blur and frequency patching, which make SSL immune to frequency-domain attacks. Lastly, we analytically explain how one generalize this defense to SL.

\subsection{Gaussian Blur}
\label{subsec:freq_aug}

Consider a clean image $x$ which, when poisoned using CTRL, becomes $\Tilde{x}$. First discrete cosine transform $\operatorname{dct}(\cdot)$ is applied to $x$ to get $x_f = \operatorname{dct}(x)$. Then, a trigger $t(x)$ is injected into the frequency-domain representation $x_f$ \footnote{We skip color-space conversion from RGB to YUV and back in our discussions, for brevity}, resulting in $\Tilde{x}_f = x_f + t(x)$. In the CTRL method, the frequency-space trigger \( t(x) \) is defined as:
\[
t_{i, j}(x) = 
\begin{cases} 
c, & \text{if } i = j \text{ and } i \in \left\{\frac{s}{2}, s \right\}, \\
0, & \text{otherwise},
\end{cases}
\]
where \( i \) and \( j \) are indices corresponding to the frequency components, \( c \) is a constant value that represents the strength of the trigger, and \( s \) is the spatial resolution of the frequency space. The trigger is injected at specific diagonal elements corresponding to mid-range and high frequencies.

The poisoned image $\Tilde{x}$ is obtained by applying the inverse discrete cosine transform $\operatorname{dct}^{-1}(\cdot)$ to $\Tilde{x}_f$. The image-space trigger $r(x)$ is defined as the residual $r(x) = \Tilde{x} - x$.


Blurring an image $x$ to get $x_b$ is effectively a convolution operation $x_b = k * x$, where $*$ denotes 2D linear convolution. When we blur a poisoned image $\Tilde{x}$ to get $\Tilde{x_b}$, 
\begin{gather}
    \begin{aligned}
    \Tilde{x_b} &= k * \Tilde{x} \\
                &= k * \operatorname{dct}^{-1}(x_f + t(x)) \\
                &= k * \operatorname{dct}^{-1}(x_f) + k*\operatorname{dct}^{-1}(t(x))\\
                &= k * x + k * \operatorname{dct}^{-1}(t(x))\\
                &= x_b + \describe{k * \operatorname{dct}^{-1}(t(x))}{trigger}
    \end{aligned}
\end{gather}

Thus the image-space trigger $r_b(x)$ after applying blur is $r_b(x) = k * \operatorname{dct}^{-1}(t(x))$ or $r_b(x) = k * r(x)$. Since $r_b(x) \ne r(x)$, the trigger is not invariant to blur augmentation, which reduces the attack success rate (ASR) by 3 folds (Table \ref{tab:res:ctrl}). Figure \ref{fig:clusters} visually shows the effectiveness of the augmentation. Note that we have used Gaussian blur instead of Gaussian noise as an augmentation since using the latter is detrimental to the model's classification accuracy.
\subsection{Frequency Patching (Freq Patch)}
\label{subsec:freq_patch}

Applying frequency-patching to an image $x$ to get $x_p$ is effectively a multiplication operation in frequency domain $\operatorname{dct}(x_p) = k \times x_f$, where $\times$ denotes element-wise multiplication and $k$ is a random mask containing a patch $\mathcal{P}$ of zeros, with ones everywhere else, such that $(0, 0)\not\in\mathcal{P}$. When we frequency-patch a poisoned image $\Tilde{x}$ to get $\Tilde{x_p}$, 
\begin{gather}
    \begin{aligned}
    \Tilde{x_p} &= \operatorname{dct}^{-1}(k\times\Tilde{x_f}) \\
                &= \operatorname{dct}^{-1}(k\times(x_f + t(x))) \\
                &= \operatorname{dct}^{-1}(k\times x_f) + \operatorname{dct}^{-1}(k\times t(x))\\
                &\approx x + \describe{\operatorname{dct}^{-1}(k\times t(x))}{trigger} \qquad \text{[appx 1]}
    \end{aligned}
\end{gather}

The above approximation is a direct result of the fact that $\operatorname{dct}(\cdot)$ concentrated most of its energy at $(0, 0)$, which is not in $\mathcal{P}$. Note that $\operatorname{dct}^{-1}(k\times t(x))$ is a complex non-linear operation in the image-space, therefore image-space trigger after frequency-patching $r_p(x) \approx \operatorname{dct}^{-1}(k\times t(x))$. This means that $r_p(x)\ne r(x)$, i.e., the trigger in CTRL is not invariant to frequency-patching augmentation. 
\subsection{Extension to Supervised Learning}
\label{subsec:extension}

\bonak{Dhruva says this looks a bit like related work / preliminary content\\}
Augmentation invariance or equivariance of a trigger pattern is critical to the success of an attack in a contrastive learning framework (Sec. \ref{subsec:equi_inv}). However, this property is absent in SL as they do not employ contrastive learning. A simple patch-based attack such as \cite{yang2020patchattack} or a single radioactive pixel attack such as \cite{radioactive} can fool deep models successfully. In \cite{robustequivariance}, the authors defend against adversarial attacks by enforcing equivariance property on neural net's ($f_\theta$) predictions w.r.t. augmentations. Specifically, they maintain a list of augmentations $\mathcal{G} = \{g_1, g_2, \dots, g_n\}$ and optimize an additional loss $\mathcal{L}_{equi}$ on top of supervised loss $\mathcal{L}_{sup}$ and regularization $\mathcal{L}_{reg}$. $\mathcal{L}_{equi}$ is defined as:
\begin{gather}
    \begin{aligned}
    & \mathcal{L}_{equi}: 
    && \sum_{g_i \in \mathcal{G}} d(g^{-1}_i \circ f_{\theta} \circ g_i(\mathbf{x}), f_{\theta}(\mathbf{x})) \\ 
    &or\hspace{2em} && \sum_{g_i \in g} d(f_{\theta} \circ g_i(\mathbf{x}), g_i \circ f_{\theta}(\mathbf{x})) 
    \end{aligned}
\end{gather}
if $g^{-1}$ is doesn't exist. The authors use image-space augmentations such as random crop, random rotation, etc. in their paper. One can additionally use frequency-space augmentations such as Gaussian blur and frequency-patching to improve robustness against frequency-space attacks. Specifically, we must maintain two lists $\mathcal{G}_{equi}$ and $\mathcal{G}_{inv}$ which contain augmentations w.r.t. which the downstream task is equivariant and invariant respectively. For example, blurring an image doesn't change segmentation masks, therefore making image segmentation task invariant w.r.t. Gaussian blur augmentation. Our new loss $\mathcal{L}_{aug}$ is defined as:
\begin{gather}
    \begin{aligned}
    & \mathcal{L}_{aug}: 
    && \sum_{g_i \in \mathcal{G}_{equi}} d(f_{\theta} \circ g_i(\mathbf{x}), g_i \circ f_{\theta}(\mathbf{x})) \\ 
    &\hspace{2em} &&  + \sum_{g_i \in \mathcal{G}_{inv}} d(f_{\theta}(\mathbf{x}), f_{\theta} \circ g_i(\mathbf{x}))
    \end{aligned}
\end{gather}
\bonak{Commenting old subsections\\}
\bonak{scrapping this as well for now}



\section{Experiments}
\label{section:experiments}

\begin{table*}[t]
    \centering
    \resizebox{\linewidth}{!}{
    \begin{tabular}{ccccccccccccccccccc}
\multicolumn{2}{c}{Method} &&& \multicolumn{3}{c}{Vanilla} && \multicolumn{3}{c}{Blur} && \multicolumn{3}{c}{Freq Patch} && \multicolumn{3}{c}{Blur + Freq Patch} \\
\cmidrule{1-2} \cmidrule{5-7} \cmidrule{9-11} \cmidrule{13-15} \cmidrule{17-19}
Arch & Dataset &&& ACC & ASR & AA &&  ACC & ASR & AA && ACC & ASR & AA && ACC & ASR & AA \\
\hline
\hline
\addlinespace[0.5em]
\multirow{ 2}{*}{SimCLR} & CIFAR 10 &&& 84.96 & 83.62 & 30.00 && \textbf{89.38} & 35.54 & 38.40 && 87.74 & 65.57 & 37.16  && 87.83 & \textbf{25.78} & \textbf{39.67}  \\
 & CIFAR 100 &&& 49.76 & 95.5  & 8.61 && 56.24 & 38.98  & 11.65 && 52.27 & 83.21 & 10.98 && \textbf{56.33} & \textbf{32.16} & \textbf{14.47}\\
\multirow{ 2}{*}{BYOL} & CIFAR 10 &&& 86.36 & 91.31 & 31.75 && \textbf{90.45} & 19.22 & \textbf{37.62} && 83.16 & 49.83 & 29.27 && 84.92 & \textbf{11.89} & 33.81 \\
 & CIFAR 100 &&& 51.71 & 92.93 & 9.67 && 59.36 & \textbf{14.96} & 15.47 && 56.29 & 83.91 & 13.37 && \textbf{60.99} & 15.85 & \textbf{16.54} \\
\multirow{ 2}{*}{SimSiam} & CIFAR 10 &&& 85.34 & 59.36 & 29.34 && \textbf{89.32} & 15.51 & \textbf{36.28} && 86.04 & 55.57 & 35.39 && 78.07 & \textbf{11.17} & 33.91 \\
 & CIFAR 100 &&& 55.08 & 86.76 & 10.27 && \textbf{60.34} & 40.63 & 13.93 && 55.17 & 77.91 & \textbf{14.28} && 57.65 & \textbf{17.85} & 13.86 \\
\addlinespace[0.2em]
\hline
\addlinespace[0.2em]
    \end{tabular}
}
    \caption{Results for three SSL models under four pre-training paradigms: (1) Vanilla augmentations, (2) Blur augmentation, (3) Frequency patching augmentation, and (4) Combination of blur and frequency patching. ACC is classification accuracy (higher is better), ASR is backdoor attack success rate (lower is better), and AA is adversarial accuracy (higher is better). Vanilla and Blur experiments ran for 800 training epochs and the other paradigms for 700 training epochs.}
    \label{tab:res:ctrl}
\end{table*}

Here, we provide details about our experimental setups, metrics used and the results. 
\bonak{write about RobustBench}
\subsection{Datasets}
We use the CIFAR 10 and CIFAR 100 datasets for all our experiments. For our backdoor defense, following \cite{CTRL}, we poison 1\% of the training set in CIFAR 10 for all attacks. Since 1\% of CIFAR 100 constitutes an entire class, we poison 0.5\% of CIFAR 100. To maintain consistency, all results shown in the paper are based on experiments carried out by setting the target class as "1".  

\subsection{Classification metrics}
For our backdoor defense evaluation, following \cite{neural_cleanse}, we use the metrics ACC (clean accuracy) to measure a backdoored classifier's performance on classifying clean samples, and ASR (Attack success rate) to evaluate the backdoor attack efficacy of the overall setup. ASR is defined as the percentage of poisoned samples that are classified as the target class. For our classifier robustness evaluation, we use a robustness measure called adversarial accuracy (AA), which is an upper bound of the fraction of data points on which the classifier predicts the correct class for all possible perturbations from a perturbation set, as described in \cite{robustbench}. 

\subsection{RobustBench}

We use RobustBench \cite{robustbench} to benchmark the adversarial robustness of our SSL models. Specifically, we use the $\ell_\infty$ perturbation attack where the threat model is defined by the set of perturbations $\Delta$ and $\Delta_\infty$ = $[\delta \in \mathbb{R}^d, \Vert \delta_{p} \Vert \leq \epsilon ]$ where $\epsilon$ is the perturbation upper bound chosen such that the ground truth label should stay the same for each in-distribution input within the perturbation set (we set $\epsilon = \frac{2}{255}$) for our implementation. For evaluation, RobustBench employs AutoAttack \cite{autoattack} which uses four attacks including a projected gradient attack (PGD) run on cross-entropy and difference-of-logits cost functions, the targeted FAB attack that aims to minimize the $\ell_p$ norm of the perturbations and the black box square attack \cite{square} that iteratively adds localized square patches at random positions as perturbations, such that each perturbation is at the boundary of the feasible perturbation set $\Delta$. Each attack is run on a distilled version of the dataset for which the previous attack has not produced adversarial or perturbed examples.  


\subsection{Results }

We present our results for our defense against the CTRL attack in Table \ref{tab:res:ctrl}. We observe that contrastive training with blur provides a two-fold benefit: ACC is notably increased while ASR is reduced by 60 - 85\% depending on the dataset and method. We also show SSL model robustness metrics with regular training, blur augmentation, frequency patching augmentation, and a combination of blur and frequency patching augmentations in Table \ref{tab:res:ctrl}. We observe that a combination of the blur and frequency patching augmentations almost always improve model robustness with respect to ASR, while just the blur augmentation produces better results than the frequency-patched and vanilla augmented models. This indicates that the blur augmentation is more akin to providing robustness, and produces the best model robustness when combined with frequency patching.



\section{Conclusion}
This work conducts a systematic study on the adversarial robustness of SSL against backdoor attacks. We reveal the fundamental understanding behind the working of a backdoor attack through the concept of variance (Sec. \ref{subsec:freq_aug}) and devise a defense strategy based on it. By proving our defense's effectiveness in SSL, we lay the foundation for robustness against backdoor attacks in other paradigms of learning, such as Supervised Learning, and we hope to achieve this in subsequent work. We hope our findings will shed light on generalized defense strategies against backdoor attacks and encourage the community to consider backdoor vulnerabilities while training, while also improving adversarial robustness of models.

\renewcommand{\paragraph}[1]{\noindent\textbf{#1}\quad}

{
    \small
    \bibliographystyle{styles/IEEEbib}
    \bibliography{main}
}




\appendix

\setcounter{section}{0}
\renewcommand\thesection{\Alph{section}}

\section{Implementation Details}
\label{appendix:impl_det}


\subsection{Encoder training}
We use the training split of our dataset to train self-supervised encoders (SimCLR and BYOL) in a contrastive fashion. We use hyperparameters set by \cite{CTRL} for the encoder training process. Table \ref{tab:hyperparameters} lists the key hyperparameters used in Encoder training. 

\bgroup
\def\arraystretch{1.3}
\begin{table}[h]
    \centering
    \begin{tabular}{lccc}
        \hline
        {Hyperparameter} & {SimCLR} & {BYOL} \\
        \hline
        {Optimizer} & SGD & SGD \\
        {Learning Rate} & 0.06 & 0.06 \\
        {Momentum} & 0.9 & 0.9 \\
        {Weight Decay} & $5 \times 10^{-4}$ & $5 \times 10^{-4}$ \\
        {Epochs} & 800 & 800 \\
        {Batch Size} & 512 & 512 \\
        {Temperature} & 0.5 & -- \\
        {Moving Average} & -- & 0.996 \\
        \hline
    \end{tabular}
    \caption{Hyperparameters for SSL Methods (SimCLR, BYOL)}
    \label{tab:hyperparameters}
\end{table}
\egroup


\subsection{Classifier training}
Following \cite{CTRL}, we use K-Nearest Neighbours on the embeddings generated by SSL encoder to report the ACC and ASR values. Since AutoAttack requires the gradient on the classification probabilities, we train a single-layer classifier to classify the embeddings SSL backbone embeddings and report AA metric.

\section{Further experiments}

\subsection{Backdoor attack performance}
We study the CTRL backdoor attack performance by varying the poison ratio and the poison trigger magnitude. We observe that poison trigger magnitude and the attack success rate (ASR) are clearly correlated (Table \ref{tab:poison_mag}), and that with a high enough magnitude the blur defense is mitigated. However the trigger with a high magnitude would also be visible to the naked eye. The relation between poisoning ratio and the ASR is less clear (Table \ref{tab:poison_ratio}) and result suggests that the model is able to learn the trigger pattern with as low as 0.5\% or 0.1\$ of the dataset poisoned for CIFAR10 and CIFAR100 respectively.

\begin{table}[t]
    \centering

    \begin{tabular}{cccccc}
\multicolumn{2}{c}{Method} &&& \multicolumn{2}{c}{Augment. ASR (\%)}  \\
\cmidrule{1-2} \cmidrule{5-6} 
Dataset & Magnitude &&& Vanilla & Blur \\
\hline
\hline
\addlinespace[0.5em]
    \multirow{ 3}{*}{CIFAR 10}  & 50 &&& 54.71 & 16.19\\
                                & 100 &&& 86.81 & 30.91\\
                                & 200 &&& 99.36 & 92.35\\
    \addlinespace[0.2em]
    \addlinespace[0.2em]
    \multirow{ 3}{*}{CIFAR 100} & 50 &&& 10.56 & 3.88 \\
                                & 100 &&& 82.06 & 28.16 \\
                                & 200 &&& 97.30 & 77.24 \\

\addlinespace[0.2em]
\hline
\addlinespace[0.2em]
    \end{tabular}
    \caption{CTRL backdoor attack performance with varying poison magnitudes using SimCLR. Posion ratio is set at 1\%}
    \label{tab:poison_mag}
\end{table}

\begin{table}[t]
    \centering

    \begin{tabular}{cccccc}
\multicolumn{2}{c}{Method} &&& \multicolumn{2}{c}{Augment. ASR (\%)}  \\
\cmidrule{1-2} \cmidrule{5-6} 
Dataset & Ratio &&& Vanilla & Blur \\
\hline
\hline
\addlinespace[0.5em]
    \multirow{ 4}{*}{CIFAR 10}  & 0.5\% &&& 89.81 & 53.34\\
                                & 1\% &&& 86.81 & 30.91\\
                                & 2\% &&& 90.08 & 47.21\\
                                & 3\% &&& 77.2 & 43.65\\
    \addlinespace[0.2em]
    \addlinespace[0.2em]
    \multirow{ 4}{*}{CIFAR 100} & 0.1\% &&& 69.43 & 32.44\\
                                & 0.2\% &&& 82.06 & 28.16\\
                                & 0.4\% &&& 83.48 & 29.18\\
                                & 0.6\% &&& 75.05 & 38.66\\

\addlinespace[0.2em]
\hline
\addlinespace[0.2em]
    \end{tabular}
    \caption{CTRL backdoor attack performance with varying poison ratios using SimCLR. Posion magnitude is set at 100}
    \label{tab:poison_ratio}
\end{table}

\subsection{Performance of clean models}
We present in Table \ref{tab:Clean_performance}, results on models trained with clean data. For the sake of consistency, we use same hyperparameters for models with and without defense. We observe that blur further enhances KNN accuracy by a noticeable amount.

\begin{table}[t]
    \centering

    \begin{tabular}{cccccc}
\multicolumn{2}{c}{Method} &&& \multicolumn{2}{c}{Augment. ACC (\%)}  \\
\cmidrule{1-2} \cmidrule{5-6} 
Arch & Dataset &&& Vanilla & Blur \\
\hline
\hline
\addlinespace[0.2em]
    \multirow{ 2}{*}{SimCLR}  & CIFAR 10 &&& 84.73 & \textbf{89.34}\\
      & CIFAR 100 &&& 48.36 & 55.94\\
    \addlinespace[0.4em]
    \multirow{ 2}{*}{BYOL} & CIFAR 10 &&& \textbf{86.19} & \textbf{90.36}\\
    & CIFAR 100 &&& \textbf{52.41} & \textbf{59.93}\\
    \addlinespace[0.4em]
    \multirow{ 2}{*}{SimSiam} & CIFAR 10 &&& 85.71 & \textbf{89.50}\\
    & CIFAR 100 &&& 53.64 & \textbf{59.25}\\
\addlinespace[0.2em]
\hline
\addlinespace[0.2em]
    \end{tabular}
    \caption{Performance of models on clean dataset}
    \label{tab:Clean_performance}
\end{table}

\section{Related work:} Backdoor attacks and defenses have been extensively studied in Deep Learning, especially in supervised settings. The goal of attacks is to cause "controlled misclassification" of a model in a way that suits the attacker.

\textbf{Backdoor Attacks:} Backdoor attacks can be described as poisoning a small subset of training data with trigger-embedded target class inputs, where a trigger is a manual perturbation introduced by the attacker to create a mapping between the trigger and the target class. A lot of work has been done designing triggers to increase backdoor efficiency. \cite{badnets} introduce a pixel-based trigger for model backdooring. \cite{targeted_backdoors} use a blended injection strategy to generate triggers, where a trigger pattern is partially added to the target image, with a magnitude such that its effect is quite potent as a backdoor, but it remains visibly obscure. Inspired from stenography, \cite{invisible_backdoors} demonstrate a trigger strategy using Least Significant Bit (LSB) substitution to create sub-pixel perturbations that are indistinguishable to the human eye but can be detected using neural networks (allowing for the trigger pattern to get associated with the target class). \cite{poisonedencoder} introduce a trigger strategy tailored for contrastive learning by concatenating multiple input images to create poisoned inputs and repeating this process iteratively to end up with a poisoned dataset such that two augmentations of poisoned images produce features similar to target class images. \cite{augmentation_backdoors} show how simple augmentations considered standard for SSL can be converted into triggers. \cite{fiba_paper} (FIBA) and \cite{CTRL} (CTRL) demonstrate highly effective backdoor attacks using frequency-domain triggers.  \\

\par
\textbf{Backdoor Defenses:}
Various backdoor defense strategies exist that aim towards detecting poisoned samples, recovering clean samples from poisoned samples, or reverse-engineering the poison to mitigate its effect on the trained encoder or the downstream task. \cite{neural_cleanse} identified the first generalizable strategy to detect as well as \emph{'un-poison'} poisoned samples. They detect poisoned samples by considering the minimum perturbation required to transform an output label to the target and reverse-engineer the trigger pattern to obtain clean samples.  
\cite{decoupling_backdoor} identify poisoned samples by differential clustering on trained backdoored models. \cite{smoothinv} regenerate the poison pattern by a two-step process where first, they build a robust classifier through a de-noising diffusion process, then iteratively synthesize the poison pattern by iteratively optimizing the cross-entropy between the target class and robust classifier outputs. \cite{backdoor_defense_regularization} use controlled layer-wise weight initialization and knowledge distillation to defend using unlabelled data. \cite{distillationdefense}, \cite{distillationdefense2} also utilize knowledge-distillation to defend against backdoored models.


\textbf{Backdoors and SSL:} SSL methods have recently been shown to be vulnerable to backdoor attacks through model access (\cite{badencoder}) or via poisoning a small subset of the training data (\cite{backdoor_attacks_umbc}). \cite{CTRL} (CTRL) demonstrates a highly successful attack against SSL. \cite{backdoor_attacks_umbc} also demonstrates a primitive defense against backdoor attacks, while \cite{backdoor_defense_umbc} displays a more comprehensive method to detect poisoned samples in SSL.

\section{Exploring other attack strategies against SSL}

\subsection{FIBA: A frequency-injection based backdoor attack}
\label{subsec:FIBA}
FIBA\cite{fiba_paper} is a frequency injection-based backdoor attack targetted against medical image segmentation. 




Fiba works by injecting the trigger image's amplitude component into an image's low-frequency region. Mathematically, given a trigger image $p$ and a target image $x$, FIBA injects a poison to obtain $\Tilde{x}$ such that:
\begin{gather}
    \begin{aligned}
    \mathcal{A}, \Theta &= \operatorname{polar}(\operatorname{fft}(x))\\
    \mathcal{A}_p, \Theta_p &= \operatorname{polar}(\operatorname{fft}(p))\\
    \Tilde{\mathcal{A}} &= \big[(1-\alpha)\mathcal{A}_p + \alpha \mathcal{A}\big] \cdot \mathcal{M} + \mathcal{A} \cdot (1 - \mathcal{M})\\
    \Tilde{x} &= \operatorname{fft}^{-1}(\Tilde{\mathcal{A}}\operatorname{exp}(i\cdot\Theta)) \\
    \end{aligned}
\end{gather}

where $\mathcal{A}, \Theta$ denote the amplitude and phase components respectively, and $\mathcal{M}$ is a low-frequency mask

\begin{equation}
\mathcal{M}_{ij} = 
\left
\{\begin{array}{c l}
1 & \quad (i,j) \in \text{low freq. patch}\\
0& \quad \text{otherwise}
\end{array}.
\right.
\end{equation}

\begin{figure}[tp]
    \centering
    \begin{minipage}[b]{0.5\textwidth}
        \centering
        \includegraphics[width=\textwidth]{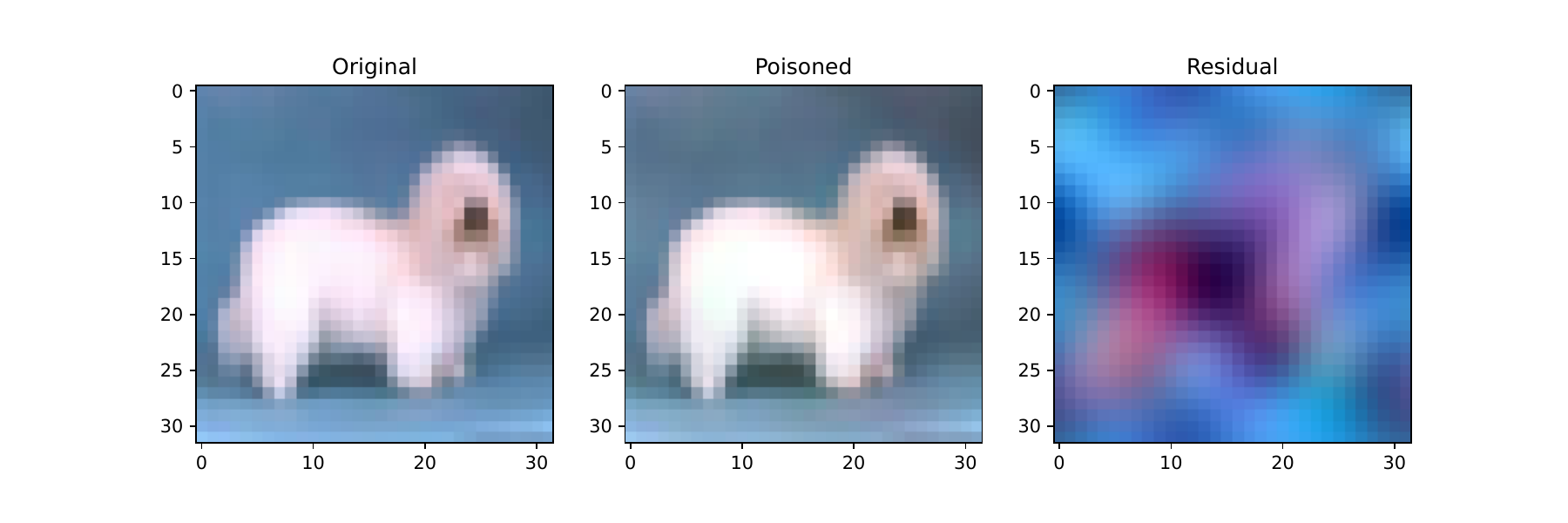}
    \end{minipage}
    \begin{minipage}[b]{0.5\textwidth}    
        \vspace{-1.5ex}
        \centering
        \includegraphics[width=\textwidth]{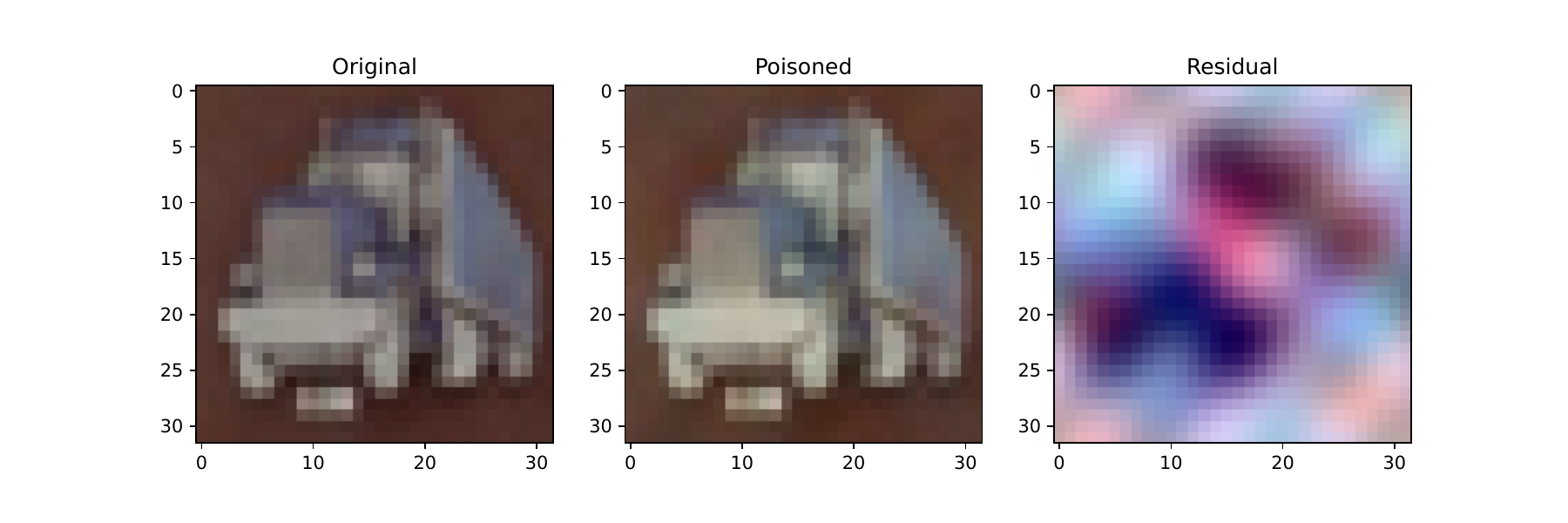}
    \end{minipage}
    \caption{Example of FIBA poisoning on two different images. The residuals, i.e. the difference between original and poisoned image vary between images. Scaled version of residuals have been plotted for visualization purposes.}
    \label{fig:fiba-residuals}
\end{figure}

The residual $t(x) = \Tilde{x} - x$ is non-linearly dependent on the target image $x$. This implies that the trigger pattern $t(x)$ has a high variance for a given target class (Fig \ref{fig:fiba-residuals}), which becomes difficult for the model to learn. This is also confirmed by our findings in Table \ref{tab:res:fiba}, where the ASR is close to random chance ($\frac{100}{\text{num class}} \%$) regardless of the presence of frequency-space augmentations.

\begin{table}[t]
    \centering
    \resizebox{\linewidth}{!}{
    \begin{tabular}{ccccccccc}
\multicolumn{2}{c}{Method} &&& \multicolumn{2}{c}{Vanilla} && \multicolumn{2}{c}{Blur}  \\
\cmidrule{1-2} \cmidrule{5-6} \cmidrule{8-9}
Arch & Dataset &&& ACC & ASR && ACC & ASR \\
\hline
\hline
\addlinespace[0.2em]
    \multirow{ 2}{*}{SimCLR}  & CIFAR10 &&& 84.96 & \textbf{9.948}  && \textbf{89.52} & 10.06\\
      & CIFAR100 &&& 48.51 & \textbf{0.906} && \textbf{56.32} & 1.046\\
    \addlinespace[0.4em]
    \multirow{ 2}{*}{BYOL} & CIFAR10 &&& 85.98 & \textbf{10.14}  && \textbf{90.32} & 10.21\\
    & CIFAR100 &&& 52.23 & \textbf{0.83} && \textbf{59.77} & 0.948\\
\addlinespace[0.2em]
\hline
\addlinespace[0.2em]
    \end{tabular}
    }
    \caption{Comparision of defense on FIBA}
    \label{tab:res:fiba}
\end{table}

\subsection{Hidden trigger backdoor attacks (HTBA): a patch-based attack strategy}
\label{subsec:htba}

HTBA \cite{backdoor_attacks_umbc} introduces small trigger patches into a small subset of target class images. They theorize that the model prefers to associate the easier-to-learn triggers with the target class rather than through a semantic understanding of the image. However, they note that their attack would be ineffective if the trigger were lost during augmentations, which commonly occurs in SSL, and we confirm this with the low ASR values in our findings in Table \ref{tab:res:htba}. We also observe that the pattern in ASR and ACC metrics are consistent with our findings in Table \ref{tab:res:ctrl}. 


\begin{table}[t]
    \centering
    \resizebox{\linewidth}{!}{
    \begin{tabular}{ccccccccc}
\multicolumn{2}{c}{Method} &&& \multicolumn{2}{c}{Vanilla} && \multicolumn{2}{c}{Blur}  \\
\cmidrule{1-2} \cmidrule{5-6} \cmidrule{8-9}
Arch & Dataset &&& ACC & ASR && ACC & ASR \\
\hline
\hline
\addlinespace[0.2em]
    \multirow{ 2}{*}{SimCLR}  & CIFAR10 &&& 84.36 &\textbf{17.3}  && \textbf{89.08} & 17.39\\
      & CIFAR100 &&& 49.03 & 16.61  && \textbf{56.35} & \textbf{12.76}\\
    \addlinespace[0.4em]
    \multirow{ 2}{*}{BYOL} & CIFAR10 &&& 85.69 & 17.59  && \textbf{90.39} & \textbf{16.41}\\
    & CIFAR100 &&& 51.71 & 22.63  && \textbf{59.85} & \textbf{18.67}\\
\addlinespace[0.2em]
\hline
\addlinespace[0.2em]
    \end{tabular}
    }
    \caption{Comparision of defense on HTBA}
    \label{tab:res:htba}
\end{table}



    \section{Limitations}

While the literature on backdoor attacks against supervised learning is abundant, very little research exists on backdoor attacks targetting SSL, owing to the adversarial robustness of SSL and the absence of labels for training. This work tests our defense against CTRL, FIBA, and HTBA. Due to time constraints, we could only evaluate our work in CIFAR 10 and CIFAR 100. We plan to conduct our experiments on other datasets (e.g. Imagenet100 which was introduced in \cite{backdoor_attacks_umbc}) and SOTA foundation models (e.g. MoCo v3, SimSiam). For our experiments, we follow all hyperparameters available from \cite{CTRL} and appropriately assign values to the remaining, which is not necessarily optimal. 
The future scope of this research includes extending the attack and the defense to multi-modal SSL regimes, e.g., CLIP. Additionally, theoretical understanding behind the working of backdoor attacks in terms of invariance and equivariance, opens room for researchers to find the optimal trigger pattern which is resistant to defenses.

\end{document}